# Hybrid Deepfake Detection Utilizing MLP and LSTM

Jacob Mallet
*Department of Computer Science*
*University of Wisconsin – Eau Claire*
Eau Claire, WI, USA
malletjc3227@uwec.edu

Natalie Krueger
*Department of Computer Science*
*University of Wisconsin – Eau Claire*
Eau Claire, WI, USA
kruegena4954@uwec.edu

Dr. Mounika Vanamala
*Department of Computer Science*
*University of Wisconsin – Eau Claire*
Eau Claire, WI, USA
vanamalam@uwec.edu

Dr. Rushit Dave
*Department of Computer Information Science*
*Minnesota State University, Mankato*
Mankato, MN, USA
rushit.dave@mnsu.edu

*Abstract*—The growing reliance of society on social media for authentic information has done nothing but increase over the past years. This has only raised the potential consequences of the spread of misinformation. One of the growing methods in popularity is to deceive users through the use of a deepfake. A deepfake is a new invention that has come with the latest technological advancements, which enables nefarious online users to replace one's face with a computer-generated, synthetic face of numerous powerful members of society. Deepfake images and videos now provide the means to mimic important political and cultural figures to spread massive amounts of false information. Models that are able to detect these deepfakes to prevent the spread of misinformation are now of tremendous necessity. In this paper, we propose a new deepfake detection schema utilizing two deep learning algorithms: long short-term memory and multilayer perceptron. We evaluate our model using a publicly available dataset named 140k Real and Fake Faces to detect images altered by a deepfake with accuracies achieved as high as 74.7%.

## I. Introduction

In the modern world, digital media has a large impact on the opinions of the public. Specifically, media that originates from well-known people such as politicians or celebrities. Deepfakes can take advantage of this impact and use it for malicious purposes. A deepfake is a digitally created photo or video of a person in which it is not really them, but an altered version of them. Deepfake technology has progressed to the point that almost anyone can easily impersonate someone else without their permission. This has allowed many people to maliciously create fake photos and videos of well-known public figures, painting them in a negative light or making it seem as if they are saying or doing something that they have not. This media can spread rapidly and cause public outrage or confusion when the deepfake is realistic enough to trick the average person. This is a prominent reason why research is needed to develop ways of detecting deepfakes accurately, helping to stop the spread of malicious media and to create a more informed public.

This paper provides a new method of deepfake image detection that uses two different machine learning algorithms. Machine learning has been shown to be effective when used for image classification [1], user authentication [2-13], and other security functions [14-16]. This is evidence that it is also an effective method for detecting deepfakes. The algorithms we have tested in this study are Long Short-Term Memory Network (LSTM) and Multilayer Perceptron (MLP). These methods have all been shown to produce accurate results when used for deepfake detection [17-19]. The dataset we are using to test these algorithms is called 140k Real and Fake Faces [20], a publicly available dataset retrieved from Kaggle. This dataset consists of 70,000 images from two different datasets, Flickr-Faces-HQ [21], which contains entirely real faces, and the Deepfake Detection Challenge dataset [22], containing deepfake faces created using style Generative Adversarial Networks (GANs). The novel contributions of this study include the testing of the two mentioned algorithms on their ability to classify real and fake images, all on the same dataset.

## II. Related Work

### A. LSTM

In one study [17], researchers use a convolutional LSTM-based residual network, CLRNet, to detect deepfakes. This method focuses on deepfake videos rather than images, detecting the inconsistencies between frames of a video. It also uses a convolutional LSTM to overcome a lack of spatial information recorded with other LSTM methods. This includes 3D tensors that record two dimensions of spatial information. Sets of five frames are taken from videos in multiple datasets, resized, and put through data augmentation methods before being evaluated by the algorithm. CLRNet is compared to several current methods on three different tests of transfer learning. The method performs best with an accuracy of 97.18% when a single source dataset and a single target dataset were used. CLRNet has shown to be a superior architecture when compared to previous baseline models, and provides a step towards improved future deepfake detection.

### B. CNN

Funding provided by University of Wisconsin-Eau Claire's Office of Research and Supported Projects

Another study [18] features a Convolutional Neural Network (CNN) used alongside a Vision Transformer (ViT) to create a Convolutional Vision Transformer method. This study also works to classify deepfake videos. First, faces are extracted from the videos and the data is augmented to prepare for classification. The next step, performed by a CNN with seventeen convolutional layers, is feature learning. This involves extracting learnable features from the selected faces. Each layer of the CNN has three to four convolutional operations, and the final product is a feature map of the inputted face images. This map is then passed through the ViT, which uses an encoder to classify the faces as real or fake. Researchers split multiple premade datasets into training, validation, and testing sets. In an experiment, thirty facial images are evaluated at a time. The proposed model reached a maximum accuracy of 93%, and performed similarly to compared models. One issue with this experiment was inherent issues in the face detection software used in preprocessing, which misclassified many images. The study concludes that the model has room for improvement in multiple ways, but the high accuracy shows that CLRNet is a good example of a generalized model.

## C. MLP

Researchers in [19] used a hybrid of a CNN and an MLP to classify deepfake videos. Both algorithms are used separately and in different ways. Initially, frames from a video are inputted into a CNN, which performs automatic feature extraction. However, before the MLP is used, data must be extracted from the frames with facial landmark detection. This is done by a pre-trained detector. The data includes a number of eye blinks and shape features such as eye and nose width and lip size. This data is normalized and passed as input to the MLP, which uses two layers to produce an initial classification. Once both the MLP and the CNN have produced output, it is mixed together and linked to an activation layer and a connected neural layer. These produce the final result, a classification of the deepfake.

An experiment is conducted using 199 fake videos and 53 real videos from a premade dataset, and 66 real videos from a separate set. The model was compared to another that only includes a CNN. Results showed that the testing accuracy of the hybrid model reached 87%, with the CNN model at 74%. The proposed model also provides a faster training period. An important note is that the CNN model overfits earlier than the proposed model, which resulted in a drop in test accuracy. This model provides a base for future work on classifying deepfakes quickly and with few computational resources.

## III. METHODOLOGY

### A. Classifier Background

This section will detail the background of the two classifiers used in this study, namely LSTM and MLP, detailed in Tables 1, 2, 3, and 4, as well as Fig. 1 and 2. Multilayer Perceptron is a deep-learning algorithm and is repeatedly used in classification contexts due to its efficiency in this domain. MLP is known as a feed-forward neural network, meaning the data flows through the model advancing from the input layer to the output layer, never retreating. Neurons named perceptrons compose the MLP algorithm, with each perceptron receiving n features, and a weight is assigned to each individual feature.

TABLE I. RESULTS FOR ALL ALGORITHMS

| Algorithms | Metrics | | | |
|---|---|---|---|---|
| | *Accuracy* | *Precision* | *Recall* | *F1 Score* |
| LSTM | 74.7% | 71.8% | 81.4% | 76.3% |
| MLP | 68% | 69% | 61% | 64% |
| CNN | 88.3% | 89.9% | 86.3% | 88.1% |
| SVM | 81.7% | 84.8% | 77.2% | 80.8% |

During the training process of an MLP classifier, a supervised learning approach is used called backpropagation. This technique attempts to minimize error within the model and trains using gradient descent. During training, the weights associated with each feature can be updated. This identifies and prioritizes the inputted features that are found to be most important by the algorithm. MLP is a versatile deep learning algorithm, in that it has the ability to handle either a linear or continuous function. This enables the algorithm to be applied in several different contexts, including image classification.

Long Short-Term Memory is another type of artificial neural network and deep learning algorithm. LSTM is very similar to a Recurrent Neural Network (RNN), in that they both have the ability to process sequences of data, in addition to individual data points like most deep learning algorithms. One major issue with traditional RNN networks stems from their failure to leverage and remember data in the long term. LSTM attempts ot solve this issue by placing more of an emphasis on remembering long-term dependencies.

During the training process of LSTM, the primary goal the network is striving for is to minimize loss as much as possible. To calculate the loss, a gradient is used, which is the loss with respect to several weights within the data. The specific weights are constantly being adjusted to produce the lowest amount of loss possible. For each time step, data flows through an input gate where the features that are identified by the network as meaningful are extracted.

The remaining data is again assessed to evaluate the long-term relevance of the data, and any features identified as such are scaled and added to the cell state. The data continues to flow into an output gate, where the cell state is determined if it is ready to make a prediction, or if the cell needs to be fed back into the LSTM network for further refinement. After traversing through the data, the model is ready to compile and make predictions.

TABLE II. AUC FOR ALL ALGORITHMS

| Algorithms | Metrics | | | |
|---|---|---|---|---|
| | *LSTM* | *MLP* | *CNN* | *SVM* |
| AUC | 74.7% | 66% | 88.3% | 81.7% |

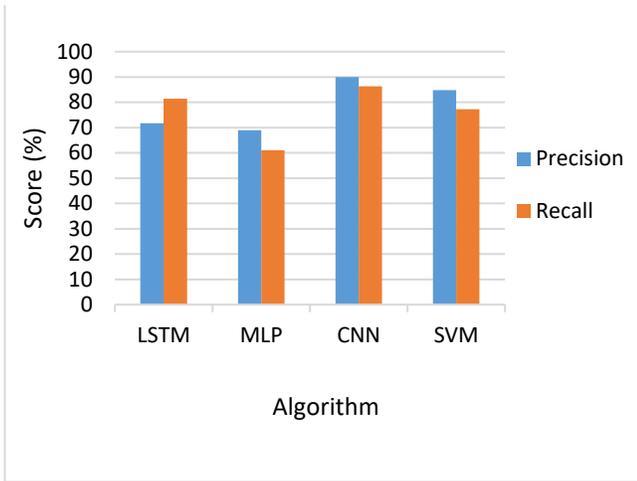

Fig. 1. Comparing Precision and Recall

### B. System Overview

We will overview both models evaluated in our research in this section, starting with the preprocessing steps taken. Both of our models use data with the same preprocessing methodology applied. First, images are read using an Image Data Generator. Grayscale, resizing, and zooming were applied to the images primarily to focus more on the person of interest's face in the image and trim out some of the background. These images are then outputted and converted to a three-dimensional array, which will be directly fed into each model.

To expand on our model based on the LSTM classifier, we simply use two layers. The first being the LSTM layer, where the activation function is set to sigmoid. The next and final layer is a dense output layer to provide the prediction for this model. Finally, we compile the model using a binary cross-entropy loss function to place the predictions into one class or the other, a deepfake or authentic image.

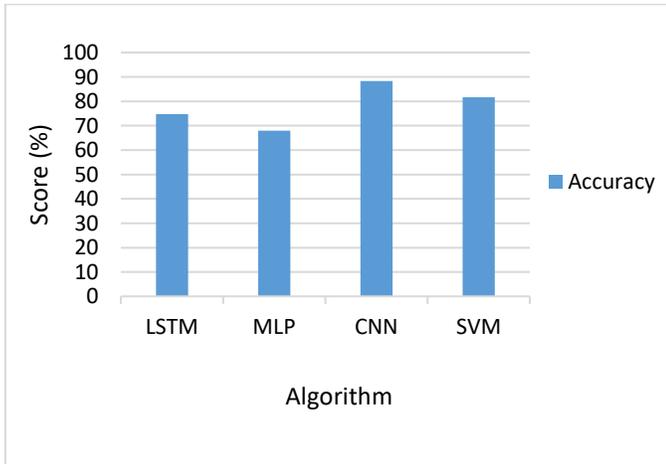

Fig. 2. Comparing Accuracies Across Algorithms

Moving on to our model using an MLP classifier, here we use four layers in total. The first layer flattens the input data further in order to advance into our fully connected dense layer, which uses a sigmoid activation function. The dense layer is essentially repeated, with the difference being half of the units used in the previous layer. After these two fully connected layers, the final layer, serving as the output layer, is a dense layer with a sigmoid activation function applied here again. The same compilation process as the previous model is applied to our MLP-based model again.

### C. Dataset

The dataset used to evaluate our various models in this study is called 140k real and fake faces [20]. This dataset merges two separate datasets together, namely Flickr-Faces-HQ [21] and the Deepfake Detection Challenge dataset [22] to create an expansive and robust set of images to benchmark our model. 70,000 images come from the Flickr-Faces-HQ dataset, with all images being of real faces of different people. Another 70,000 images are taken from the Deepfake Detection Challenge dataset, which contains deepfakes generated by a style GAN [20]. This style GAN used holds the ability to learn authentic facial features and patterns, only then to apply that knowledge in generating deepfakes. Using a dataset that contains an equivalent amount of deepfakes and real faces is essential for training our model without biasing in one direction. We split these images into a 75/25 train-test split. All of the images are preprocessed using the steps detailed earlier and fed to an image data generator to obtain an output capable of being learned by the neural networks. Within the image data generator process, the images are zoomed in centrally by 20% to isolate the face within the image further.

### D. Ethical Disclaimer

On top of the safety concerns regarding misinformation that come with the use of deepfakes, research in this area comes with an abundance of ethical issues in itself. When using a deepfake, whether with malicious intent or not, a real person's face is used as the source of the generation of the deepfake, unless the face is entirely computer generated. A key aspect of the data used in this study is the satisfaction of the latter scenario, meaning all our deepfakes are completely computer-generated faces. Therefore, none of the synthetic faces created were done so with the intent of taking a real person's face and applying their likeness to someone else. In addition to the deepfakes included in this study, 70,000 real faces are used as well. These images containing authentic faces were captured with consent from the users. On no occasion was there the use of synthetic or real faces without the permission granted from the participants that allowed for their likeness to be used in the creation of their respective datasets.

TABLE III. CONFUSION MATRIX FOR LSTM

|  | *Real* | *Deepfake* |
|---|---|---|
| Real | 11,895 | 5,605 |
| Deepfake | 3,261 | 14,239 |

TABLE IV. CONFUSION MATRIX FOR MLP

|  | *Real* | *Deepfake* |
|---|---|---|
| Real | 12,642 | 4,858 |
| Deepfake | 6,876 | 10,624 |

*E. Performance Metrics*

To measure the performance of the two models, we use six success metrics, namely accuracy, precision, recall, F1 Score, Area Under the Curve (AUC), and a confusion matrix. Perhaps the most widely used metric, accuracy, is evaluating the percentage of time the model simply correctly identifies if the face is a deepfake or real. The second metric recorded in this study is precision. Precision gives us an idea of the rate at that our model is correctly identifying positives, or a real face in our case. That is, precision looks at all of the positive predictions and identifies the percentage that were actually real faces. Recall also deals with positive predictions, but accounts for the rate that real images were correctly identified. This time, all the real faces are gathered, and the rate that the real faces were correctly classified as real faces represents the recall. Recall can also be referred to as the true positive rate. The third metric we use, F1 Score, is a balance between precision and recall. Another metric used as a benchmark, the Area Under the Curve, is the area underneath the receiving operator characteristic (ROC) curve. The ROC curve plots a curve representing the TPR, or recall, and the false positive rate (FPR). FPR identifies the rate at which a deepfake is incorrectly identified as a real face. Again, the AUC is the area underneath the curve that results from these two parameters. The final success metric used to evaluate our models is a confusion matrix. A confusion matrix has four quadrants in a two by two table. The upper left quadrant of the matrix represents the number of true positives from our model, similar to recall. The upper right quadrant displays the number of false positives predicted, similar to the FPR. The lower left quadrant represents the number of false negatives. This is the number of times the model predicted that a real image was a deepfake. The final, lower right quadrant tallies the number of the true negatives. This would be the model successfully identifying a deepfake in our use case.

IV. RESULTS AND DISCUSSION

Many conclusions can be drawn by analyzing the results produced by the deep learning algorithms used in this study. In [1], Convolutional Neural Network (CNN) and Support Vector Machine (SVM) algorithms are used on the same dataset. These results can be used as a benchmark for the algorithms used in this study. A comprehensive view of all metrics across all algorithms can be seen in Table 1. One of the most significant conclusions found within our results is the performance of SVM and especially CNN, compared to MLP and LSTM. CNN outperformed MLP and LSTM in every metric used to evaluate the algorithms.

As for the models developed in this study, LSTM yielded slightly higher results in each metric observed in comparison to MLP. Using LSTM and MLP resulted in accuracies of 74.7% and 68% respectively. A comparison of accuracies between the four algorithms evaluated can be seen in Figure 2. Regarding precision, both algorithms produced very similar results with LSTM at 71.8% and MLP at 69%. One big disparity found between the two algorithms was in the sensitivity, or recall. Recall was the lowest metric yielded by MLP at 61%, while it was the highest for LSTM at 81.4%. This 20.4% difference shows a great contrast in the two algorithms' ability to identify true positives, meaning when LSTM is given a real face, it is able to classify the image correctly at a much higher rate than MLP. An in-depth comparison of recall and precision for each algorithm can be seen in Figure 1.

When looking at the confusion matrix, MLP actually has a higher raw number of true positives, but the number of false negatives is over double the total observed in LSTM's confusion matrix, which provides further context to the recall disparity observed. A lower recall in comparison to precision, like MLP exhibits, would indicate the model is classifying too many real images as deepfakes. LSTM has the opposite issue, where the model has a lower precision compared to recall, thus the algorithm is incorrectly classifying deepfakes as real images at a higher rate. LSTM produced a higher F1 Score than MLP, with the algorithms producing scores of 76.3% and 64% respectively.

When a curve is graphed to represent the true and false positives, we achieved an AUC of 74.7% for LSTM and 66% for MLP. These AUCs show the ability our models have to effectively distinguish between deepfake and real images. Finally, our confusion matrices can be seen in Table 3 and Table 4. As noted earlier, the most glaring difference between the two algorithms can be seen in the bottom two quadrants displaying the false negatives and true negatives. This can lead us to conclude LSTM is more proficient in correctly identifying deepfakes in comparison to MLP. The two algorithms perform similarly for true and false positive results, but the biggest takeaway is LSTM's superior ability to classify deepfakes compared to MLP.

V. LIMITATIONS

One limitation that is a part of this study is the costs of using this model in the real world. To detect deepfakes in the wild, every image would have to be tested. This amount of testing would be extremely costly with the number of deepfakes being created and distributed every day. Another limitation that frequently affects deepfake detection studies is the generalizability and transferability of a model. When a model is trained on a single dataset, it can be difficult to apply the model to real-world situations or use it on other datasets without a major drop in performance. This limitation is also applicable to this research. Using a dataset for training that contains a diverse collection of deepfakes made using multiple methods is something that could minimize this problem in future studies.

VI. CONCLUSION & FUTURE WORK

Through this paper, we have introduced a new method of deepfake detection for use with images. We used a dataset consisting of both real and fake images, 140k real and fake faces. We tested two different algorithms, Long Short-Term Memory Network and Multilayer Perceptron. Through these models, we achieved an accuracy of up to 74.7% with the LSTM algorithm. Although these results are a good indicator that this model can accurately detect deepfake images, there is still a need to make more progress if it is to be used in real-world situations. It was previously stated that a model trained on only one dataset is likely to experience a drop in performance in the real world. Increasing the ability of our model to move towards realistic applications is a necessary next step. We can accomplish this by training and testing our method on different, more robust datasets to create a smoother transition into real-world use. We also plan to test our model with other algorithms to find the best performing machine learning algorithms possible.

ACKNOWLEDGMENT

Funding for this project has been provided by the University of Wisconsin-Eau Claire's Office of Research and Special Programs Summer Research Experience Grant